\documentclass[conference]{IEEEtran}
\IEEEoverridecommandlockouts
\usepackage{cite}
\usepackage{amsmath,amssymb,amsfonts}
\usepackage{algorithmic}
\usepackage{graphicx}
\usepackage{textcomp}
\usepackage{xcolor}
\usepackage{url}
\usepackage{booktabs}
\usepackage{lipsum}
\usepackage{float}
\usepackage[pdfborder={0 0 0}]{hyperref}
\usepackage[normalem]{ulem}
\def\BibTeX{{\rm B\kern-.05em{\sc i\kern-.025em b}\kern-.08em
    T\kern-.1667em\lower.7ex\hbox{E}\kern-.125emX}}
\begin{document}

\title{EviRCOD: Evidence-Guided Probabilistic Decoding for Referring Camouflaged Object Detection}


\author{
Ye Wang$^{a, \dagger}$ \quad
Kai Huang$^{b, \dagger}$ \quad
Sumin Shen$^{a}$ \quad
Chenyang Ma$^{c, *}$%
\thanks{$\dagger$ Equal contribution (ditto@wchscu.cn, hkd20088@163.com).\par* Corresponding author (chenyangma0311@gmail.com).}
\\[4pt]
$^{a}$West China Hospital, Sichuan University \\
$^{b}$Independent Researcher \\
$^{c}$Department of Data Science and Artificial Intelligence, Auckland University of Technology
}

\maketitle

\begin{abstract}
Referring Camouflaged Object Detection (Ref-COD) focuses on segmenting specific camouflaged targets in a query image using category-aligned references. Despite recent advances, existing methods struggle with reference-target semantic alignment, explicit uncertainty modeling, and robust boundary preservation.
To address these issues, we propose EviRCOD, an integrated framework consisting of three core components: (1) a Reference-Guided Deformable Encoder (RGDE) that employs hierarchical reference-driven modulation and multi-scale deformable aggregation to inject semantic priors and align cross-scale representations; (2) an Uncertainty-Aware Evidential Decoder (UAED) that incorporates Dirichlet evidence estimation into hierarchical decoding to model uncertainty and propagate confidence across scales; and (3) a Boundary-Aware Refinement Module (BARM) that selectively enhances ambiguous boundaries by exploiting low-level edge cues and prediction confidence. Experiments on the Ref-COD benchmark demonstrate that EviRCOD achieves state-of-the-art detection performance while providing well-calibrated uncertainty estimates. Code is available at: https://github.com/blueecoffee/EviRCOD.
\end{abstract}


\section{Introduction}
\label{sec:intro}

Camouflaged Object Detection (COD) is a challenging visual perception task that aims to identify and segment objects that are visually indistinguishable from their surroundings due to strong similarities in texture, color, and illumination. Such intrinsic blending makes COD substantially more difficult than conventional salient object detection, as the visual cues separating the camouflaged object from the background are extremely weak or even imperceptible to both humans and algorithms. Recent advances in deep learning have led to notable performance improvements in COD \cite{zhong2024survey}; however, most existing methods rely solely on information from a single image, making them highly sensitive to the degree of camouflage and background complexity. To alleviate the inherent ambiguity of single-image COD, Referring Camouflaged Object Detection (Ref-COD) has recently been introduced \cite{zhang2025referring}. Unlike conventional COD, Ref-COD incorporates a query image together with reference images from the same category, which provide category-specific semantic priors to facilitate the identification of the concealed target.
\begin{figure}[htbp]
\centering
\includegraphics[width=\columnwidth]{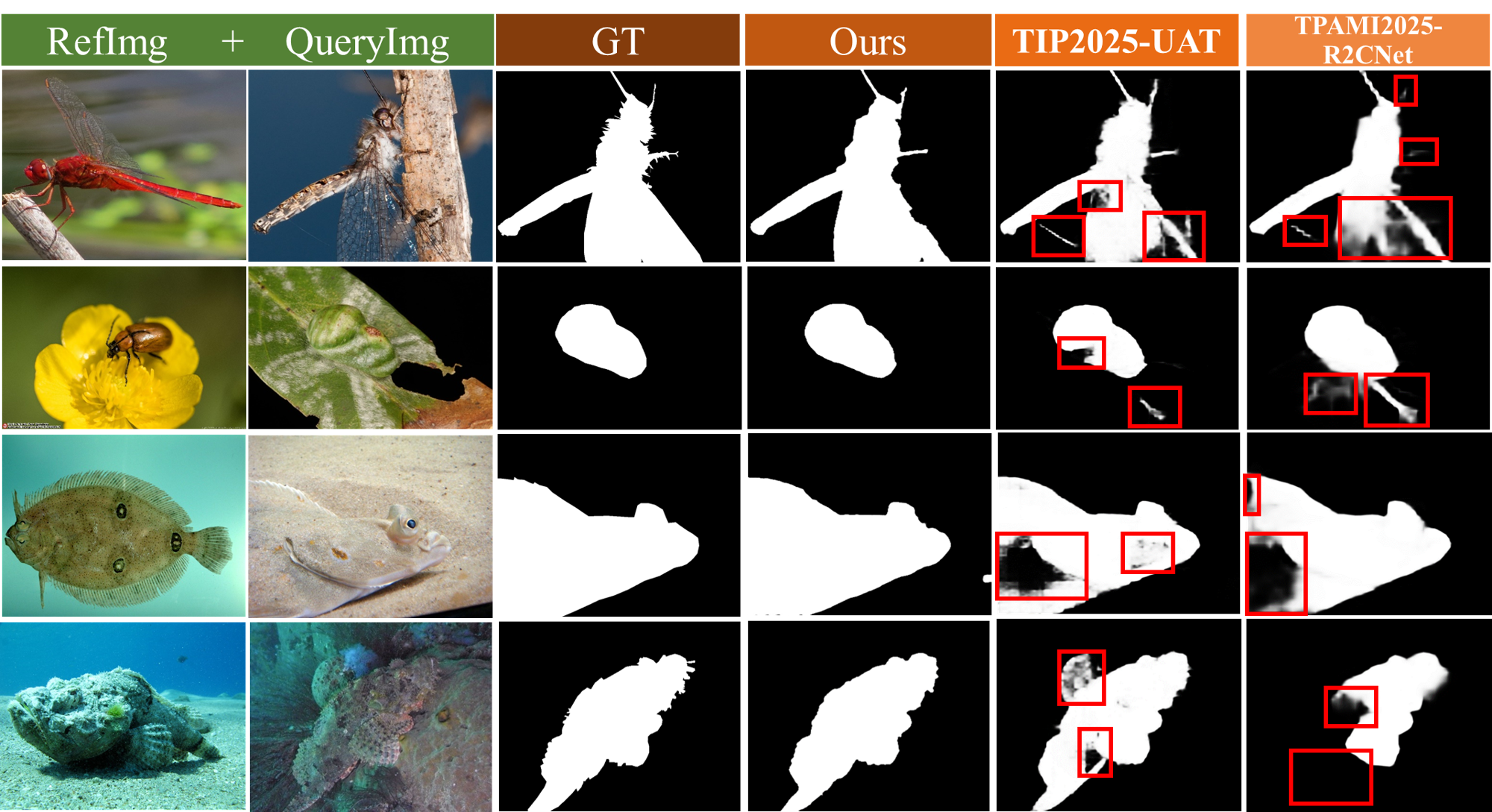}
\caption{Visual comparisons of different methods. The proposed EviRCOD generates more complete segmentation structures, sharper boundaries, and fewer background misclassifications compared to R2CNet\cite{zhang2025referring} and UAT\cite{wu2025uncertainty}.}
\label{fig1}
\end{figure}

The pioneering R2CNet \cite{zhang2025referring} established the first Ref-COD benchmark (R2C7K) and introduced reference-guided feature modulation to achieve cross-image semantic alignment. 
Building upon this, UAT \cite{wu2025uncertainty} models each token with a Gaussian distribution and derives uncertainty from the variance of multiple reparameterized samples, improving robustness to ambiguous shapes and subtle boundaries. These studies demonstrate the promise of leveraging visual references and uncertainty modeling to alleviate the inherent ambiguity in camouflaged scenes.
Nevertheless, Ref-COD remains in its early stage, and existing approaches still struggle to deliver reliable and fine-grained predictions under challenging conditions. As illustrated in Fig.~\ref{fig1}, existing approaches tend to produce fragmented predictions, coarse boundaries, or intrusion of the background. 

We identify three core challenges: (a) Insufficient semantic transfer. Current models typically fuse reference features in a coarse, fixed manner, resulting in weak cross-scale semantic alignment. This limitation often leads to incomplete segmentation, particularly within interior regions and near boundary transitions. (b) Rigid multi-scale encoding. Fixed feature aggregation schemes lack the flexibility to adapt to spatially irregular camouflage patterns, leading to degraded structural perception and imprecise boundary localization. (c) Uncalibrated uncertainty in decoding. Existing decoders generally lack reliable uncertainty modeling, which can cause background confusion and unstable boundary predictions in visually ambiguous scenes. 
These limitations underscore the need for a unified framework that (i) effectively transfers reference semantics across scales, (ii) adaptively encodes spatially varying camouflage contexts, and (iii) provides reliable and well-calibrated uncertainty estimation to guide prediction stability and boundary refinement. 

To bridge this gap, we present EviRCOD, the first Ref-COD framework that systematically introduces evidential learning \cite{sensoy2018evidential}. Unlike existing approaches that rely on stochastic sampling or ensemble-based strategies for uncertainty estimation, EviRCOD adopts a deterministic and theory-grounded evidential formulation, directly modeling prediction confidence through evidence accumulation. 
The proposed framework is designed to realize four essential capabilities: reference-guided semantic transfer, deformable context aggregation, evidence-based uncertainty quantification, and boundary-aware refinement.
These capabilities are instantiated in a unified end-to-end architecture comprising three key components:
\begin{itemize}
    \item \textbf{Reference-Guided Deformable Encoder (RGDE)} couples hierarchical reference modulation with deformable multi-scale fusion, injecting semantic priors while adaptively aligning spatially variant tokens across resolutions.
    \item \textbf{Uncertainty-Aware Evidential Decoder (UAED)} incorporates Dirichlet-based evidence theory into hierarchical decoding to jointly model epistemic and aleatoric uncertainty, enabling robust confidence propagation and improved structural consistency.
    \item \textbf{Boundary-Aware Refinement Module (BARM)} integrates fine-grained edge cues with confidence-guided gating to selectively refine ambiguous regions, achieving precise boundary recovery.
\end{itemize}

\section{Method}
\subsection{Overall Framework}
\begin{figure*}[!ht]
\centering
\includegraphics[width=\textwidth]{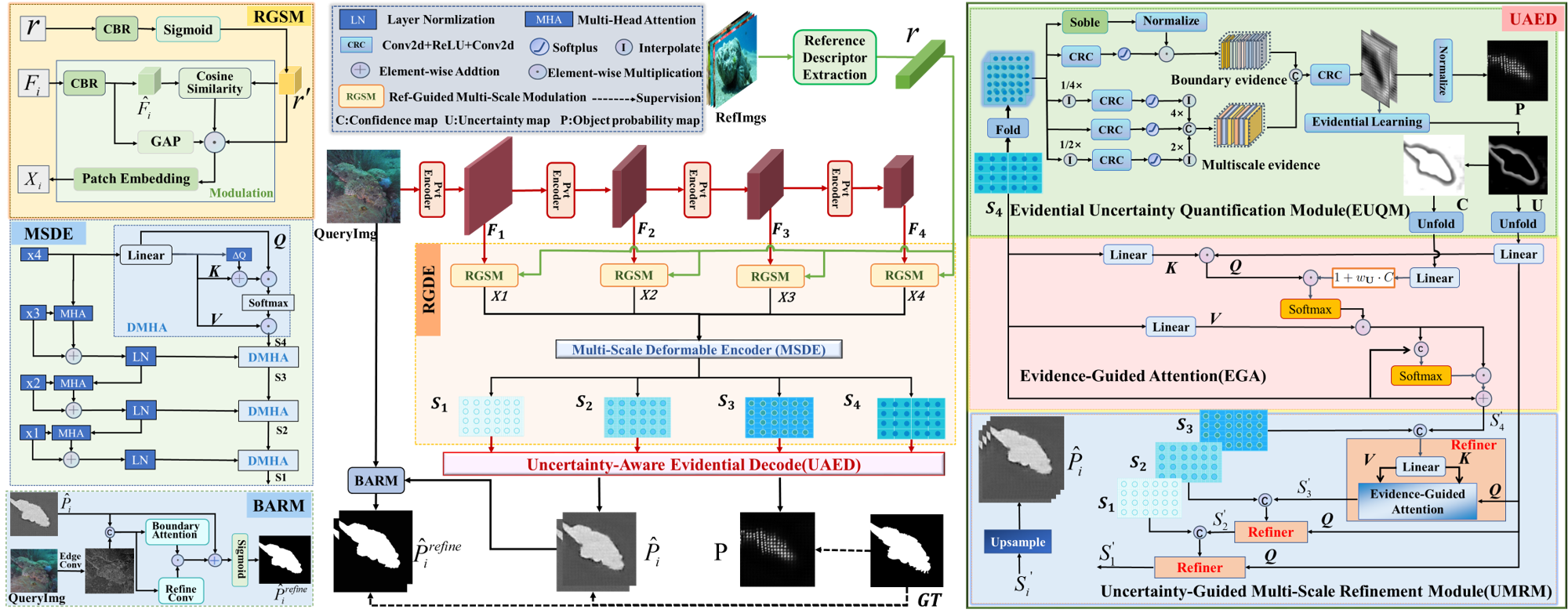}
\caption{Architecture of EviRCOD, illustrating the Reference-Guided Deformable Encoder, the Uncertainty-Aware Evidential Decoder, and the Boundary-Aware Refinement Module for uncertainty-aware Referring Camouflaged Object Detection.}
\label{fig2}
\end{figure*}

As illustrated in Fig.~\ref{fig2}, EviRCOD takes a query image and multiple reference images as input and produces a refined camouflaged object mask. A pre-trained SOD network~\cite{zhuge2022salient} extracts a compact reference descriptor providing stable category-consistent semantic priors. Guided by this descriptor, the RGDE adaptively modulates hierarchical query features and performs deformable cross-scale aggregation for fine-grained semantic alignment across resolutions. Based on the aligned representation, the UAED produces coarse predictions while modeling pixel-wise evidential uncertainty to support robust decoding under camouflage ambiguity. Finally, the BARM integrates edge cues and uncertainty-driven boundary attention to refine ambiguous contours and suppress background intrusion. Together, these components form an end-to-end framework for accurate and uncertainty-aware Ref-COD.

\subsection{Reference-Guided Deformable Encoder}
\label{sec:RGDE}

The RGDE aims to alleviate category-level semantic mismatch between query and reference images while reducing spatial and scale inconsistencies inherent in camouflaged scenes. It integrates reference-driven feature modulation with deformable multi-scale aggregation to construct a semantically aligned latent representation for robust target localization. Given a query image $\mathbf{x}$, the PVTv2 backbone~\cite{10897534} extracts four-level features $\{F_i\}_{i=1}^{4}$, where $F_i \in \mathbb{R}^{B \times C_i \times \frac{H}{2^{i+2}} \times \frac{W}{2^{i+2}}}$.

\paragraph{Reference-Guided Multi-Scale Modulation}
The reference descriptor $\mathbf{r}$ is transformed into a normalized global prior via a composite operation consisting of a $1 \times 1$ convolution, batch normalization, and ReLU activation(denoted as $\text{CBR}(\cdot)$), and sigmoid $\sigma$ to inject category-level semantic priors.
\begin{equation}
\mathbf{r}' = \sigma \big( \text{CBR}(\mathbf{r}) \big) \in [0,1]^{B \times C_d \times 1 \times 1},
\end{equation}
Each query feature is channel-aligned via $\hat{F}_i = \mathrm{CBR}_{C_d}(F_i) \in \mathbb{R}^{B \times C_d \times H_i \times W_i}$. 
To achieve scale-adaptive modulation, we compute a scale-aware modulation weight:
\begin{equation}
W_i = \mathcal{C}\big(\sigma(Q(\hat{F}_i) \odot S(\hat{F}_i, \mathbf{r}'))\big) \in [\tau_{\min}, 1]^{B \times C_d \times 1 \times 1},
\end{equation}
where $\odot$ denotes element-wise multiplication, $Q(\hat F_i)=\mathrm{GAP}(\hat F_i)\in\mathbb R^{B\times C_d\times1\times1}$,  $S(\hat{F}_i, \mathbf{r}')_c = \frac{\langle Q(\hat{F}_i)_c, \mathbf{r}'_c \rangle}{\|Q(\hat{F}_i)_c\| \cdot \|\mathbf{r}'_c\|}, \quad c = 1, \ldots, C_d.$ $Q(\hat F_i)$ provides a global statistical summary of the query features, and $S(\hat{F}_i, \mathbf{r}')$ computes channel-wise cosine similarity between the query summary and the reference prior. The clipping operator $\mathcal{C}(\cdot)$ enforces $W_i \in [\tau_{\min},1]$ to prevent excessive suppression and avoid feature collapse.
The final modulated features are given by:
\begin{equation}
\tilde{F}_i = \hat{F}_i \odot W_i \odot \mathbf{r}', \quad i=1,\ldots,4.
\end{equation}
where $\hat{F}_i$ preserves local structure, $W_i$ provides scale-adaptive weighting, and $\mathbf{r}'$ injects category-level semantic priors, operating at complementary semantic levels.
Then, all modulated features are upsampled to $88\times 88$ and tokenized into $4\times 4$ patch embeddings via a patch embedding layer $E_p$:
\begin{equation}
X_i = E_p(\tilde{F}_i) \in \mathbb{R}^{B\times N\times D},
\end{equation}
with $N = (88/4)^2 = 484$ and $D=1024$.  

\paragraph{Multi-Scale Deformable Encoder}
We adopt a top-down cross-scale attention scheme to efficiently propagate coarse semantic priors to finer scales, formulated as:
\begin{equation}
\begin{split}
\hat{X}_i = \mathrm{LayerNorm}\!\Big( &
X_i + \mathrm{MHA}(Q=X_i, \\
& K=X_{i+1}, V=X_{i+1}) \Big), i=1,2,3.
\end{split}
\end{equation}

Each $\hat{X}_i$ is further processed by a deformable encoder $\mathcal{E}_i$ based on Deformable Multi-Head Attention (DMHA)~\cite{zhu2020deformable}.  
Spatial offsets are predicted independently for each scale:
\begin{equation}
\Delta Q = f_\theta(Q) \in \mathbb{R}^{B \times N \times h \times 2},
\end{equation}
where $h$ denotes the number of attention heads. Keys are dynamically shifted as $K' = K + \gamma \Delta Q$ with $\gamma=0.1$. While coarser scales employ standard self-attention, finer scales incorporate high-level semantic masks to emphasize target regions. The encoded representations are:
\begin{align}
S_i &= \mathcal{E}_i(\hat{X}_i), i = 1,2,3, \quad S_4 = \mathcal{E}_4(X_4).
\end{align}
Through top-down semantic propagation and deformable aggregation, fine-scale representations are progressively refined with high-level semantic guidance, effectively alleviating cross-scale misalignment in camouflaged scenes.

\subsection{Uncertainty-Aware Evidential Decoder}
\paragraph{Evidential Uncertainty Quantification Module}
Decoding begins with the coarsest encoder feature $S_4$, which is reshaped into spatial form
and fed into an evidential prediction module. This module employs two parallel evidence extraction branches that integrate multi-scale semantic cues and Sobel-based boundary
information, producing a non-negative evidence tensor
$\mathbf{E} = [e_0, e_1] \in \mathbb{R}^{B \times 2 \times H' \times W'}$,
where $e_0$ and $e_1$ denote the evidence supporting the background and target classes,
respectively. Following evidential learning theory, the Dirichlet concentration parameters are then obtained as:
\begin{equation}
\boldsymbol{\alpha} = \mathbf{E} + 1, \qquad
\boldsymbol{\alpha} = [\alpha_0, \alpha_1], \qquad
S = \alpha_0 + \alpha_1,
\end{equation}
where $\alpha_0$ and $\alpha_1$ denote the evidence for the background and target classes, respectively,
and $S$ denotes the Dirichlet strength. The target-class probability
$P$ and a unified uncertainty measure $\mathcal{U}$ as:
\begin{equation}
P = \frac{\alpha_1}{S}, \qquad
\mathcal{U} = \lambda_1 \frac{2}{S}
+ \lambda_2 \frac{\alpha_1 (S - \alpha_1)}{S^2 (S + 1)}.
\end{equation}
Here, $\mathcal{U}$ combines \emph{vacuity} (epistemic uncertainty arising from insufficient evidence) and \emph{predictive variance} (aleatoric uncertainty caused by inherent data ambiguity). The weighting coefficients $\lambda_1$ and $\lambda_2$ balance these complementary components to form a task-driven uncertainty representation suitable for Ref-COD.
A lightweight $\mathrm{MLP}$ then projects $\mathcal{U}$ into an uncertainty embedding:
\begin{equation}
U = \mathrm{MLP}(\mathcal{U}).
\end{equation}

\begin{figure*}[!ht]
\centering
\includegraphics[width=\textwidth]{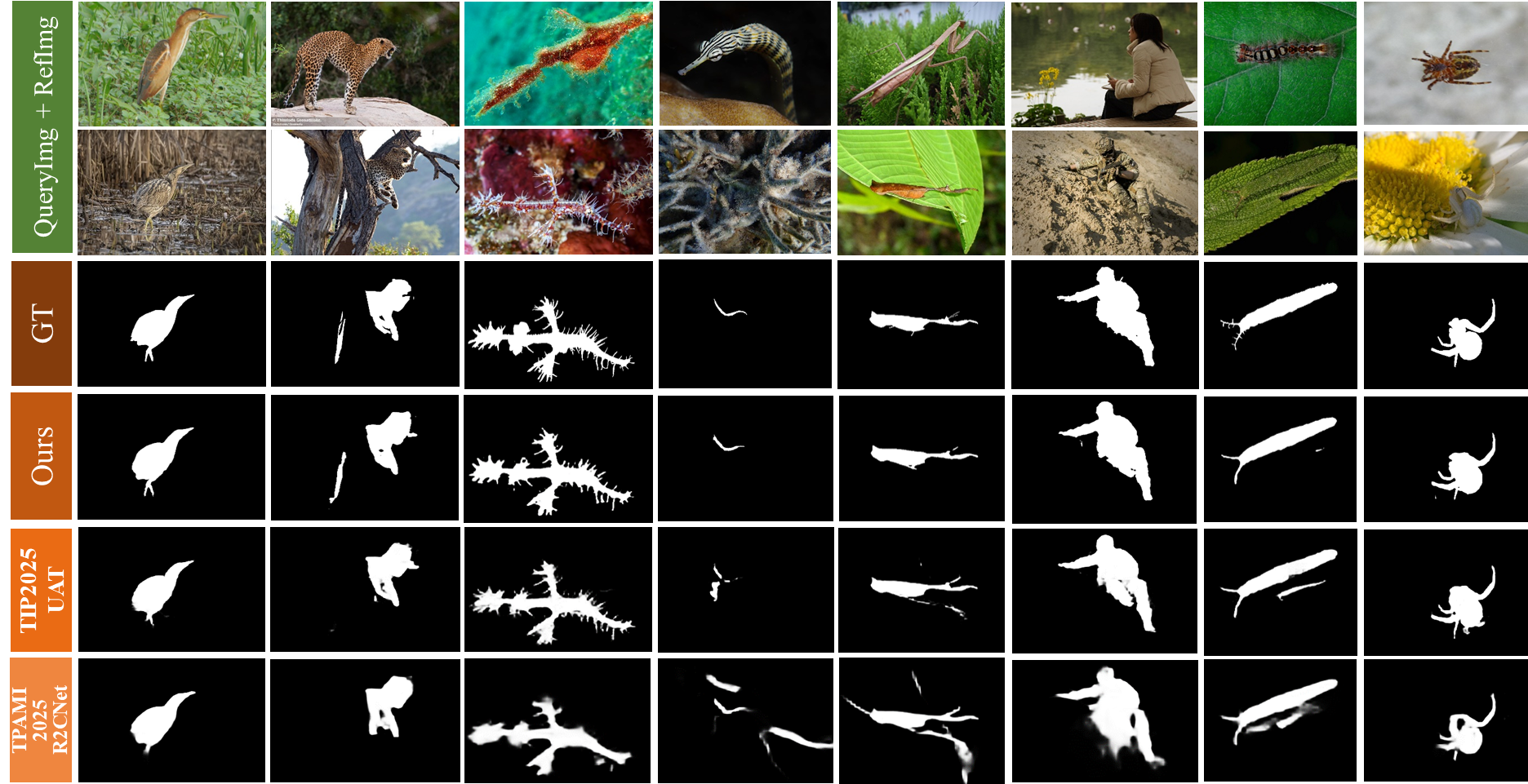}
\caption{Qualitative comparisons between EviRCOD and the main Ref-COD baselines (R2CNet\cite{zhang2025referring} and UAT\cite{wu2025uncertainty}). EviRCOD produces more coherent object masks with sharper boundaries and significantly fewer background errors, highlighting its superiority in challenging camouflage scenarios.}
\label{fig3}
\end{figure*}

\paragraph{Evidence-Guided Attention}
To enable uncertainty-aware feature refinement, we introduce an Evidence-Guided Attention
mechanism that modulates attention scores using a confidence map $C = 1 - \mathcal{U}$, such that spatial regions with higher confidence contribute more
strongly to the attention response. The attention weights are computed as:
\begin{equation}
A = \mathrm{Softmax}\!\left(
\frac{Q_{\mathbf{U}} K^{\top}}{\sqrt{D}}
\odot \big(1 + w_{\mathbf{U}} \cdot C\big)
\right), 
M = A \cdot V,
\end{equation}
where $Q_{\mathbf{U}}$ is obtained by linearly projecting the uncertainty embedding
$\mathbf{U}$, while $K$ and $V$ are linear projections of the fused features at the current decoding stage.  
The learnable coefficient $w_{\mathbf{U}}$ controls the influence of the confidence map $C$ on the attention logits, enabling confidence-aware rescaling of attention logits. $D$ denotes the feature dimension, and broadcasting is applied to ensure spatial alignment between $C$ and the attention logits. $M$ denotes the uncertainty-guided attention output obtained by aggregating the value features weighted by the attention map $A$.
We further refine the coarsest-scale feature using a gated residual formulation:
\begin{equation}
S_4' = S_4 + \sigma\!\big(G(\mathrm{Concat}(S_4, M))\big) \odot M,
\end{equation}
where $\sigma(\cdot)$ is softmax activation, $\mathrm{Concat}(\cdot,\cdot)$
denotes channel-wise concatenation, and $G(\cdot)$ is a linear gating unit.

\paragraph{Uncertainty-Guided Multi-Scale Refinement Module}
Building upon $S_4'$, we progressively refine features across finer decoder scales
($i=3,2,1$). At each refinement stage, the current-scale encoder feature $S_i$ is
concatenated with the refined output from the immediately coarser scale $S_{i+1}'$
to form a fused representation:
\begin{equation}
\tilde{S}_i = \mathrm{Concat}(S_i, S_{i+1}').
\end{equation}
The fused feature $\tilde{S}_i$ is then refined using the same Evidence-Guided
Attention to produce $S_i'$. At each refinement stage, a scale-specific prediction
map is generated as
\begin{equation}
\hat{P}_i = f_{\mathrm{up}}(S_i'),
\end{equation}
where $f_{\mathrm{up}}(\cdot)$ denotes bilinear upsampling that resizes the
prediction to the input image resolution.

\subsection{Boundary-Aware Refinement Module}
\label{sec:BAR} 
Although the uncertainty-aware decoder produces semantically coherent predictions, object boundaries often remain ambiguous under camouflage; to address this, we propose the BARM, which jointly leverages low-level edge priors and prediction confidence.
Given the decoded prediction map $\hat{P}_i$ and the query image $\mathbf{x}$, the BARM operates in three stages:

\paragraph{Edge Prior Extraction}
A lightweight edge extractor $f_{\text{edge}}$ produces a single-channel edge prior:
\begin{equation}
Edge = f_{\text{edge}}(\mathbf{x}),
\end{equation}
which captures fine-grained structural cues complementary to $\hat{P}_i$ and is spatially aligned for subsequent fusion.

\paragraph{Dual-Branch Synergistic Optimization}
The concatenated feature $[\hat{P}_i, Edge]$ is fed into two parallel branches:
\begin{itemize}
    \item \textit{Attention branch}:  
    A boundary-aware weight map $G_i = \sigma(f_{\text{attn}}([\hat{P}_i, Edge]))$ highlights boundary-uncertain regions where refinement is necessary.
    \item \textit{Refinement branch}:  
    A residual correction map $\Delta\hat{P}_i = f_{\text{ref}}([\hat{P}_i, Edge])$ models signed adjustments for boundary deviations (both inward and outward corrections).
\end{itemize}

\paragraph{Selective Boundary Refinement}
The final refined prediction is obtained via gated residual aggregation:
\begin{equation}
\hat{P}_i^{\text{refine}}
= \sigma\!\left(\hat{P}_i + G_i \odot \Delta\hat{P}_i\right),
\end{equation}
where the gating mask $G_i$ enables selective refinement of ambiguous regions while preserving confident structures, and $\hat{P}_1^{\text{refine}}$ (for $i=1$) serves as the model's final prediction map.

\begin{table*}[t]
\centering
\setlength{\tabcolsep}{3pt} 
\caption{QUANTITATIVE COMPARISON OF EviRCOD WITH STATE-OF-THE-ART METHODS. The best three results are highlighted with \textbf{bold} (1st), \underline{solid underline} (2nd), and \dashuline{dashed underline} (3rd). ``$\uparrow$'': the higher the better, ``$\downarrow$'': the lower the better.}
\label{tab:quantitative_comparison}
\begin{tabular}{l|l|cccc|cccc|cccc}
\toprule[1.2pt] 
\multicolumn{1}{c|}{Models} & \multicolumn{1}{c|}{Pub\&Year} & \multicolumn{4}{c|}{Overall} & \multicolumn{4}{c|}{Single Object} & \multicolumn{4}{c}{Multiple Objects} \\
\cline{3-14} 
 &  & $S_m\uparrow$ & $\alpha_E\uparrow$ & $\omega_ F\uparrow$ & $\mathcal{M}\downarrow$ & $S_m\uparrow$ & $\alpha_E\uparrow$ & $\omega_F\uparrow$ & $\mathcal{M}\downarrow$ & $S_m\uparrow$ & $\alpha_ E\uparrow$ & $\omega_F\uparrow$ & $\mathcal{M}\downarrow$ \\
\hline
PFNet-Ref & PFNet\cite{mei2021camouflaged}: CVPR-2021 & 0.811 & 0.885 & 0.687 & 0.036 & 0.815 & 0.886 & 0.691 & 0.035 & 0.764 & 0.873 & 0.632 & 0.045 \\
PreyNet-Ref & PreyNet\cite{zhang2022preynet}: ACMMM-2022 & 0.817 & 0.900 & 0.704 & 0.032 & 0.822 & 0.900 & 0.709 & 0.032 & 0.763 & 0.898 & 0.645 & 0.041 \\
SINetV2-Ref & SINetV2\cite{fan2021concealed}: TPAMI-2022 & 0.823 & 0.888 & 0.700 & 0.033 & 0.828 & 0.889 & 0.705 & 0.032 & 0.771 & 0.874 & 0.634 & 0.043 \\
BGNet-Ref & BGNet\cite{sun2022boundary}: IJCAI-2022 & 0.840 & 0.909 & 0.738 & 0.029 & 0.844 & 0.910 & 0.742 & 0.029 & 0.792 & 0.887 & 0.679 & \dashuline{0.036} \\ 
BSANet-Ref & BSANet\cite{Zhu2022ICF}: AAAI-2022 & 0.830 & \dashuline{0.912} & 0.727 & 0.030 & 0.827 & 0.913 & 0.733 & 0.030 & 0.774 & 0.895 & 0.655 & 0.039 \\ 
ZoomNet-Ref & ZoomNet\cite{pang2022zoom}: CVPR-2022 & 0.834 & 0.886 & 0.720 & 0.029 & 0.839 & 0.887 & 0.726 & 0.029 & 0.781 & 0.876 & 0.652 & 0.038 \\
DGNet-Ref & DGNet\cite{ji2023deep}: MIR-2023 & 0.821 & 0.891 & 0.696 & 0.032 & 0.827 & 0.890 & 0.703 & 0.031 & 0.748 & 0.879 & 0.607 & 0.045 \\
ACUMEN-Ref & ACUMEN\cite{zhang2024unlocking}: ECCV-2024 & 0.770 & 0.853 & 0.628 & 0.045 & 0.777 & 0.856 & 0.638 & 0.044 & 0.674 & 0.815 & 0.589 & 0.056 \\
GLCONet-Ref & GLCONet\cite{sun2024glconet}: TNNLS-2024 & 0.850 & 0.905 & 0.747 & \dashuline{0.027} & \underline{0.859} & \dashuline{0.913} & 0.754 & \underline{0.025} & \dashuline{0.799} & \underline{0.908} & \dashuline{0.700} & \textbf{0.032} \\ 
FSEL-Ref & FSEL\cite{sun2024frequency}: ECCV-2024 & \dashuline{0.851} & 0.906 & \dashuline{0.751} & \dashuline{0.027} & \dashuline{0.856} & 0.907 & \dashuline{0.756} & \dashuline{0.026} & 0.798 & 0.890 & 0.688 & \dashuline{0.036} \\ 
\hline
R2CNet & TPAMI-2025 & 0.805 & 0.879 & 0.669 & 0.036 & 0.810 & 0.880 & 0.674 & 0.035 & 0.747 & 0.872 & 0.602 & 0.046 \\
UAT-Res2Net-50 & TIP-2025 & 0.825 & 0.886 & 0.698 & 0.033 & 0.829 & 0.884 & 0.702 & 0.032 & 0.762 & 0.862 & 0.613 & 0.044 \\
UAT-PVTv2 & TIP-2025 & \underline{0.855} & \dashuline{0.912} & \underline{0.757} & \underline{0.026} & \underline{0.859} & \dashuline{0.913} & \underline{0.761} & \underline{0.025} & \underline{0.805} & \dashuline{0.900} & \underline{0.701} & \underline{0.033} \\ 
\hline
EviRCOD-ResNet-50 & \multicolumn{1}{c|}{-} & 0.810 & 0.903 & 0.700 & 0.033 & 0.815 & 0.905 & 0.707 & 0.032 & 0.751 & 0.882 & 0.620 & 0.043 \\
EviRCOD-Res2Net-50 & \multicolumn{1}{c|}{-} & 0.836 & \underline{0.920} & 0.741 & 0.028 & 0.840 & \underline{0.923} & 0.747 & 0.027 & 0.771 & 0.889 & 0.654 & 0.041 \\ 
EviRCOD-PVTv2 & \multicolumn{1}{c|}{-} & \textbf{0.869} & \textbf{0.944} & \textbf{0.799} & \textbf{0.021} & \textbf{0.873} & \textbf{0.946} & \textbf{0.804} & \textbf{0.020} & \textbf{0.818} & \textbf{0.920} & \textbf{0.723} & \textbf{0.032} \\
\bottomrule[1.2pt] 
\end{tabular}
\label{table1}
\end{table*}

\subsection{Joint Hybrid Loss Design}
\label{sec:JHL}
To jointly enforce spatial precision and uncertainty calibration in Ref-COD, we introduce a Joint Hybrid Loss that integrates three complementary objectives: (1) Dirichlet-based evidential supervision, (2) boundary-aware weighting for ambiguous contours, and (3) focal regularization for hard examples. In evidential learning~\cite{sensoy2018evidential}, the expected negative log-likelihood for the ground-truth class $y$ is expressed as $\psi(S) - \psi(\alpha_y)$, where $S$ denotes the Dirichlet strength and $\psi(\cdot)$ denotes the digamma function. To emphasize inherently uncertain boundary pixels in camouflaged scenes, we construct a Sobel-based weighting map $w_{\text{bnd}}$. The evidential loss is therefore defined as:
\begin{equation}
\mathcal{L}_{\text{evid}}
=
\mathbb{E}\!\left[
w_{\text{bnd}} \cdot \big(\psi(S) - \psi(\alpha_y)\big)
\right]
+
\lambda\, \mathcal{L}_{\text{focal}},
\end{equation}
where the expectation is taken over all pixels, $\mathcal{L}_{\text{focal}}$ regularizes hard or low-confidence regions and encourages the model to accumulate stronger evidence.

To further ensure regional coherence and boundary integrity, we adopt the structural loss $\mathcal{L}_{\text{str}}$ from F$^3$Net~\cite{wei2020f3net}, which combines weighted binary cross-entropy and IoU constraints. Structural supervision is applied to selected multi-scale decoder outputs $\hat{P}_i$ and refined predictions $\hat{P}_i^\text{refine}$ from the BARM. The final loss function is given by:
\begin{equation}
\mathcal{L}_{\text{total}}
=
\sum_{i=1}^{2} \omega_i \mathcal{L}_{\text{str}}(\hat{P}_i, GT)
+
\sum_{j=1}^{2} \eta_j \mathcal{L}_{\text{str}}(\hat{P}_i^\text{refine}, GT)
+
\kappa\mathcal{L}_{\text{evid}},
\end{equation}
where $\omega_i$, $\eta_j$ and $\kappa$ are scale-balancing coefficients.  
This hybrid objective jointly enforces structural fidelity, boundary sharpness, and well-calibrated uncertainty, leading to robust camouflaged object detection.

\begin{table}[t]
\centering
\caption{Ablation on R2C7K: module contributions to structure, alignment, contour, and pixel-wise metrics.}

\label{tab:ablation}
\setlength{\tabcolsep}{4pt}
\renewcommand{\arraystretch}{1.05}
\begin{tabular}{lcccc}
\toprule
\textbf{Configuration} & $S_m\uparrow$ & $\alpha_E\uparrow$ & $\omega_F\uparrow$ & $\mathcal{M}\downarrow$ \\
\midrule
Baseline & 0.838 & 0.908 & 0.751 & 0.027 \\
Baseline + RGDE & 0.854 & 0.928 & 0.775 & 0.023 \\
Baseline + UAED & 0.859 & 0.932 & 0.780  & 0.024 \\
Baseline + BARM & 0.848 & 0.917 & 0.760 & 0.025 \\
Baseline + RGDE + UAED & 0.865 & 0.940 & 0.794 & 0.022 \\
Baseline + RGDE + BARM & 0.860 & 0.933 & 0.781 & 0.022 \\
Baseline + UAED + BARM & 0.863 & 0.936 & 0.788 & 0.023 \\
\textbf{Ours} & \textbf{0.869} & \textbf{0.944} & \textbf{0.799} & \textbf{0.021} \\
\bottomrule
\end{tabular}
\end{table}

\begin{table}[t]
\centering
\caption{Ablation of UAED sub-modules (EUQM, EGA, and UMRM).}
\label{tab:uaed_ablation}
\setlength{\tabcolsep}{4pt}
\renewcommand{\arraystretch}{1.05}
\begin{tabular}{lcccc}
\toprule
\textbf{Configuration} & $S_m\uparrow$ & $\alpha_E\uparrow$ & $\omega_F\uparrow$ & $\mathcal{M}\downarrow$ \\
\midrule
EUQM & 0.867 & 0.940 & 0.793 & 0.021 \\
EGA & 0.864 & 0.937 & 0.790 & 0.021 \\
UMRM & 0.865 & 0.938 & 0.791 & 0.021 \\
EUQM+EGA & 0.867 & 0.941 & 0.796 & 0.021 \\
EUQM+UMRM & 0.868 & 0.942 & 0.797 & 0.021 \\
EGA+UMRM & 0.865 & 0.939 & 0.792 & 0.021 \\
EUQM+EGA+UMRM & \textbf{0.869} & \textbf{0.944} & \textbf{0.799} & \textbf{0.021} \\
\bottomrule
\end{tabular}
\end{table}

\section{Experiments}
\subsection{Dataset and Evaluation Metrics}
Experiments are conducted on the R2C7K dataset~\cite{zhang2025referring}, which consists of the Camo and Ref subsets, and our model is evaluated using four widely adopted metrics: Structure measure ($S_m$)~\cite{fan2017structure}, Weighted F-measure ($\omega _F$)~\cite{margolin2014evaluate}, Adaptive E-measure ($\alpha _E$)~\cite{fan2018enhanced}, and Mean Absolute Error ($\mathcal{M}$).

\subsection{Implementation Details} 
We adopt PVTv2~\cite{10897534} as the default backbone and also evaluate ResNet-50 and Res2Net-50 to demonstrate the generalizability of our framework. All models are trained end-to-end for 150 epochs on a single NVIDIA RTX 4090 GPU with a batch size of 16, using the Adam optimizer with an initial learning rate of $1\times10^{-4}$. A differential learning rate is applied: 0.1$\times$ for the backbone and 1$\times$ for the decoder and task-specific heads, decayed via cosine annealing ($T_{\text{max}}=150$) for baseline consistency.




\subsection{Comparison with State-of-the-Art Methods}
\label{sec:CM}

Ref-COD is an emerging task with two representative methods, R2CNet~\cite{fan2017structure} and UAT~\cite{wu2025uncertainty}, both of which fall short of ground-truth quality. We evaluate EviRCOD with different backbones and compare it with these Ref-COD methods as well as several conventional COD-based models adapted to the Ref-COD task. As shown in Table~\ref{table1}, EviRCOD consistently outperforms all compared methods across all metrics. Specifically, EviRCOD outperforms R2CNet and UAT under all backbone settings, and the PVTv2 variant achieves the best overall results. Qualitative comparisons in Fig.~\ref{fig3} further demonstrate more complete object coverage, sharper boundaries, better detail preservation, and improved background suppression.

\subsection{Ablation Study}
\label{sec:AS}

To validate the contributions of each component, we conduct ablation experiments by progressively adding modules to the baseline. Table~\ref{tab:ablation} reports ablation results on R2C7K. The results reveal the contribution of each component in EviRCOD, summarized as follows:
(a) Models incorporating RGDE consistently improve $S_m$ and reduce $\mathcal{M}$, indicating enhanced global structural consistency and lower pixel-wise errors;  
(b) UAED mainly improves $\alpha_E$, reflecting better alignment between predictions and ground truth and uncertainty-aware confidence modeling, with limited impact on $\mathcal{M}$;  
(c) BARM yields consistent gains in $\omega_F$ and moderate improvements in $S_m$, validating its role in boundary refinement while minimally affecting $\mathcal{M}$. 
(d) Combining all modules achieves the best overall performance, demonstrating complementary effects of global structure modeling, uncertainty-aware decoding, and boundary refinement.

Table~\ref{tab:uaed_ablation} shows that the full integration of UAED (EUQM, EGA, and UMRM) achieves the best overall performance, confirming their complementary roles in uncertainty-aware decoding. Moreover, our method achieves competitive efficiency, with 276.7M parameters and an average inference time of 37.4 ms per image (26.8 FPS), indicating a favorable trade-off between accuracy and speed.

\subsection{Uncertainty Quantification Analysis}
\begin{figure}[H]
	\centering
	\includegraphics[width=\columnwidth]{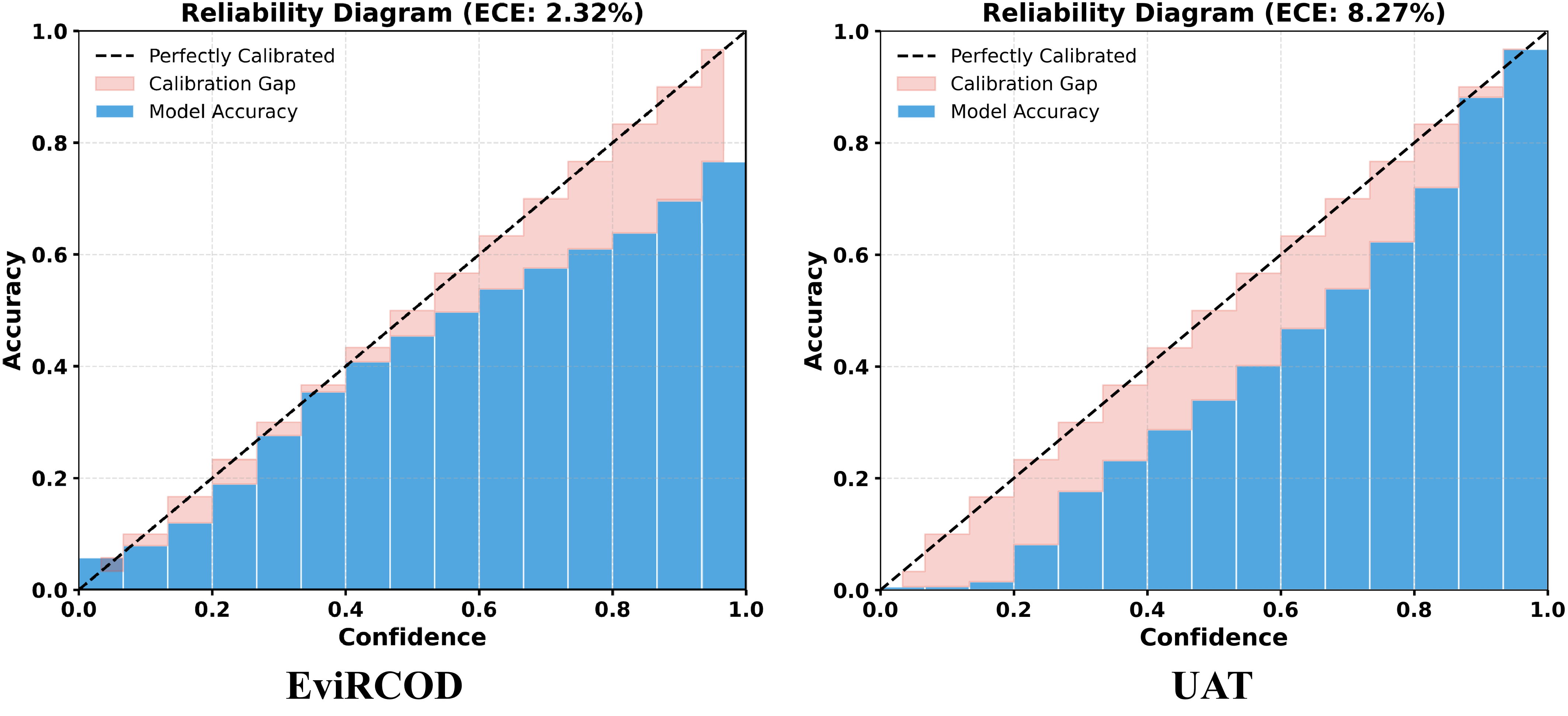} 
	\caption{Calibration comparison between UAT and our method.}
	\label{fig4}
\end{figure}

We evaluate uncertainty calibration using Reliability Diagrams and the Expected Calibration Error (ECE)~\cite{posocco2021estimating}, which quantify the consistency between predictive confidence and empirical accuracy. As shown in Fig.~\ref{fig4}, our method achieves a lower ECE (2.32\%) than UAT (8.27\%). While UAT exhibits pronounced over-confidence in nearly all confidence bins, our method demonstrates significantly improved alignment with the identity line across the confidence spectrum, indicating more reliable and better-calibrated uncertainty estimates.

\section{Conclusion}
In this paper, we present EviRCOD, a unified framework for Ref-COD that integrates reference-guided feature alignment, evidential uncertainty modeling, and boundary-aware refinement. Experiments on R2C7K show that our method outperforms state-of-the-art approaches across all standard metrics. In addition to accurate predictions, EviRCOD provides interpretable uncertainty estimates that highlight ambiguous regions, offering both performance and reliability. This work highlights the potential of uncertainty-aware learning in fine-grained and ambiguous visual perception tasks.

\bibliographystyle{IEEEbib}
\bibliography{icme2026references}

\end{document}